\documentclass[letterpaper]{article} 
\usepackage{aaai24}  
\usepackage{times}  
\usepackage{helvet}  
\usepackage{courier}  
\usepackage[hyphens]{url}  
\usepackage{graphicx} 
\usepackage{natbib}  
\usepackage{caption} 
\usepackage{algorithm}
\usepackage{algorithmic}
\usepackage{bm}
\usepackage{amsmath}
\usepackage{amssymb}
\usepackage{color}
\usepackage{booktabs}
\usepackage{multirow}
\usepackage{subfig}
\usepackage{newfloat}
\usepackage{listings}
\DeclareCaptionStyle{ruled}{labelfont=normalfont,labelsep=colon,strut=off} 
\floatstyle{ruled}
\newfloat{listing}{tb}{lst}{}
\floatname{listing}{Listing}
\title{LMD: Faster Image Reconstruction with Latent Masking Diffusion}
\author{
    Zhiyuan Ma\textsuperscript{\rm 1}, Zhihuan Yu\textsuperscript{\rm 2}, Jianjun Li\textsuperscript{\rm 2}\thanks{Corresponding authors.}, Bowen Zhou\textsuperscript{\rm 1}\footnotemark[1]
}
\affiliations{
    \textsuperscript{\rm 1}Department of Electronic Engineering, Tsinghua University\\

    \textsuperscript{\rm 2}School of Computer Science and Technology, Huazhong University of Science and Technology

    mzyth@tsinghua.edu.cn
}

\begin{document}

\maketitle

\begin{abstract}
As a class of fruitful approaches, diffusion probabilistic models (DPMs) have shown excellent advantages in high-resolution image reconstruction. On the other hand, masked autoencoders (MAEs), as popular self-supervised vision learners, have demonstrated simpler and more effective image reconstruction and transfer capabilities on downstream tasks. However, they all require extremely high training costs, either due to inherent high temporal-dependence (i.e., excessively long diffusion steps) or due to artificially low spatial-dependence (i.e., human-formulated high mask ratio, such as 0.75). To the end, this paper presents LMD, a simple but faster image reconstruction framework with \underline{\textbf{L}}atent \underline{\textbf{M}}asking \underline{\textbf{D}}iffusion. First, we propose to project and reconstruct images in latent space through a pre-trained variational autoencoder, which is theoretically more efficient than in the pixel-based space. Then, we combine the advantages of MAEs and DPMs to design a progressive masking diffusion model, which gradually increases the masking proportion by three different schedulers and reconstructs the latent features from simple to difficult, without sequentially performing denoising diffusion as in DPMs or using fixed high masking ratio as in MAEs, so as to alleviate the high training time-consumption predicament. Our approach allows for learning high-capacity models and accelerate their training (by 3$\times$ or more) and barely reduces the original accuracy. Inference speed in downstream tasks also significantly outperforms the previous approaches\footnote{Code is available: https://github.com/AnonymousPony/lmd.}.
\end{abstract}

\section{Introduction}
\begin{figure}[t]
	\centering
	\includegraphics[width=\linewidth,height=1.0\linewidth]{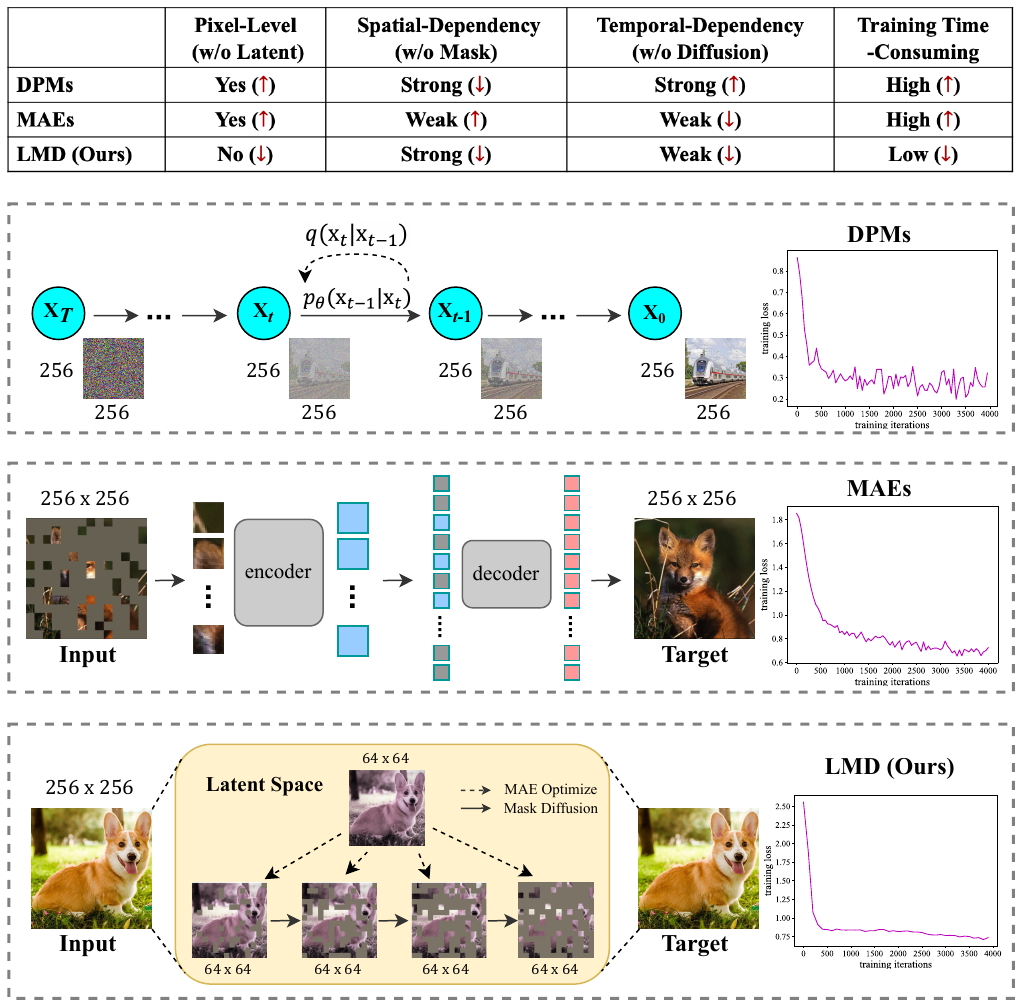}
        \vspace{-6pt}
	\caption{An example to illustrate the comparisons between our proposed LMD versus DPMs and MAEs on training time-consumption.}
	\label{fig: example}
	\vspace{-6pt}
\end{figure}

Image reconstruction is one of the most challenging and computationally expensive tasks in computer vision. Recent years, diffusion probabilistic models (DPMs)~\cite{ho2020denoising} based on Markov probability prediction have gained prominence in various generative models~\cite{rombach2022high,saharia2022photorealistic,ma2023follow}, due to their excellent performance in the diversity and high-fidelity of image synthesis. Subsequently, a large number of works based on DPMs have been proposed and great progresses have been achieved in terms of sampling procedure~\cite{liu2022pseudo}, conditional guidance~\cite{nichol2021glide}, likelihood maximization~\cite{kim2022maximum} and generalization ability~\cite{gu2022vector}. However, almost all existing DPMs face an inherent time-consuming dilemma (e.g., \emph{150-1000 V100 days in ADM}~\cite{dhariwal2021diffusion}), because the Markov diffusion process in DPMs requires a very large number of diffusion steps (e.g., \emph{thousands}) and sampling time, which can be optimized but is basically inevitable for effective synthesis~\cite{cao2022survey,croitoru2022diffusion}. Meanwhile, masked image modeling, as a simpler and more efficient self-supervised learning technique, has been introduced to the CV fields to develop a widespread Vision Transformer (ViT)~\cite{dosovitskiy2020image}, which is a non-autoregressive milestone model and has recently sparked the interest of many researchers~\cite{zhang2022hivit,ma2022simvtp}. With the great success of ViT, a series of masked autoencoders (MAEs) have been proposed~\cite{he2022masked,wei2022masked,li2022semmae}. These MAEs are widely viewed as efficient visual learners, and can fully utilize the parallel computing capability of GPU to learn multi-head visual features and reconstruct images well at pixel-level. 
Compared with DPMs, MAEs seem to be more acceptable, due to their higher inference efficiency, wider generalization performance~\cite{ma2023hybridprompt} and lower theoretical threshold. However, they also seem to be somewhat simple and crude by mandatory setting the mask proportion as a fixed number (e.g., $75\%$), which makes the training of MAEs less elegant as compared to the progressive ways in DPMs. A large number of experiments~\cite{huang2022green,liu2022mixmim,chen2022efficient} also prove that the training of MAEs is very time-consuming~(e.g., \emph{112 V100 days for MAE}~\cite{he2022masked}), which is even close to that of DPMs.

To sum up, DPMs and MAEs have made significant contributions in their respective technical routes for image reconstruction. However, they both share a fatal problem,  \emph{high training time-consumption}. Through an in-depth analysis of these two categories of models, we attribute this problem to two factors:
(1) \emph{Pixel-level modeling}. \textbf{Firstly}, they almost all model image features at the pixel-level, which naturally requires a lot of time to encode (or add-noise) and decode (or de-noise) the image features in the entire pixel space. Taking the 256$\times$256 image input in Figure~\ref{fig: example} as an example, for DPMs, each diffusion step requires Gaussian sampling and noise prediction in the 256$\times$256 pixel space, which greatly reduces its training efficiency. 
For MAEs, the pixel size of its input image determines the number of patches (i.e., \emph{the input length of the ViT}) when the same patch size is maintained. 
Therefore, for both DPMs and MAEs, modeling at pixel-level is one of the significant causes of high time-consumption. (2) \emph{Spatial-temporal dependency}. \textbf{Secondly}, their training efficiency is implicitly limited by their inherent or artificially spatial-temporal dependency. As illustrated in the table in Figure~\ref{fig: example}, for DPMs, each Markov diffusion step is calculated based on the previous  step, so the training time theoretically is positively correlated with its temporal-dependency, that is, stronger time dependence (i.e., \emph{longer diffusion steps}) will lead to higher training time-consumption. On the contrary, the training time seems to be negatively correlated with the spatial-dependency, that is, in each training iteration, the de-noising optimization process from $\textbf{X}_{t}$ to $\textbf{X}_{t-1}$ is obviously easier and the time-consumption is shorter than that directly from $\textbf{X}_t$ to $\textbf{X}_0$. Therefore, for MAEs, each iteration from 75\% masked $\textbf{X}_t$ to unmasked $\textbf{X}_0$ will naturally require higher time-consumption for training.

Driven by the above two aspects, we propose a simple but well-considered latent masking diffusion framework for faster image reconstruction. Specifically, to address the first factor, inspired by the success of Stable Diffusion, we employ a pre-trained variational autoencoder to compress the input image from the original pixel space to a latent space with smaller scale for latent destruction and reconstruction. But different from Stable Diffusion, we then follow the MAEs' architecture to split the latent feature map into small patches to compose a patch sequence with length $l$ as the readout. Based on the latent patch sequence, we further address the second factor by designing a progressive masking diffusion strategy, which gradually increases the masking proportion by mask schedulers and restores the latent features from simple to difficult. 
Among them, three different schedulers are adopted to avoid temporal-dependency for faster parallel computing and enhance spatial-dependency to help model reconstruct efficiently, so as to ultimately reduce the total training time-consumption.

Experiments on two representative datasets ImageNet-1K and Lsun-Bedrooms demonstrate the effectiveness of the proposed LMD model, showing that it achieves competitive performance against previous DPMs or MAEs models, but with significantly lower mean training time-consumption. The inference speed of LMD in image reconstruction also significantly outperforms the previous approaches. Moreover, LMD can be well generalized to a variety of downstream tasks, due to its flexible architecture.

\begin{figure*}[t]
	\centering
	\includegraphics[width=1\linewidth]{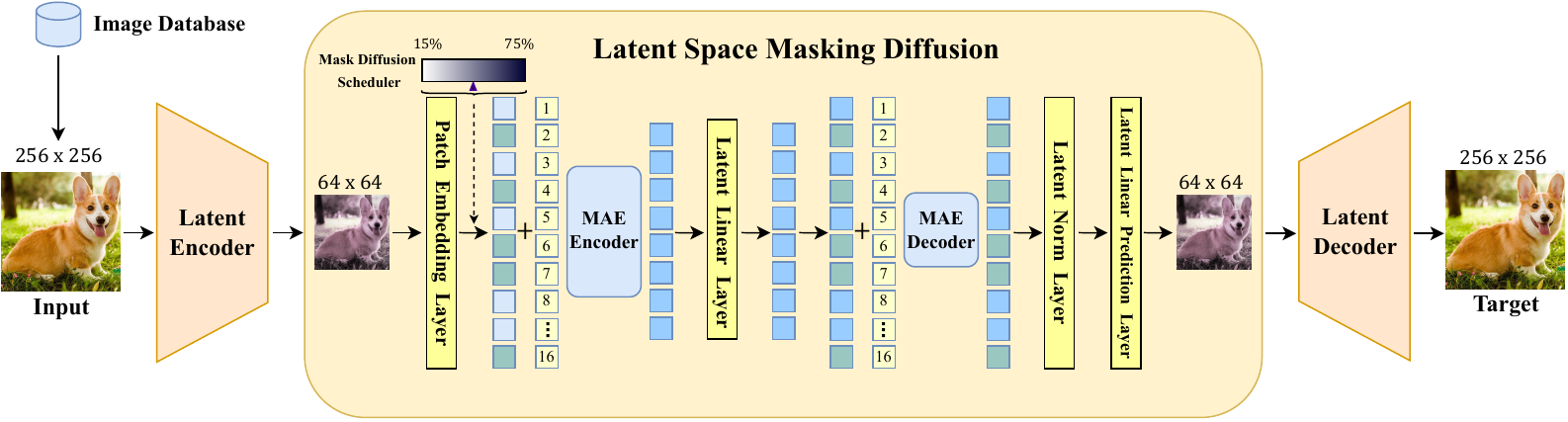}
        \vspace{-15pt}
	\caption{The Proposed Framework.}
	\label{fig: model}
	\vspace{-10pt}
\end{figure*}

\section{Related Work}
\noindent\textbf{Diffusion Probabilistic Models (DPMs).} Recent years has witnessed the remarkable success of DPMs, due to its impressive generative capabilities. After surpassing GAN on image synthesis~\cite{dhariwal2021diffusion}, diffusion models have shown a promising algorithm and emerged as the new state-of-the-art among the deep generative models~\cite{yang2022diffusion}. As a pioneering work, \textbf{DDPM}~\cite{ho2020denoising} still suffers from slow sampling procedure and sub-optimal log-likelihood estimations. To this end, \textbf{DDIM}~\cite{song2020denoising} proposes a more efficient sampling procedure to accelerate the forward process and has been widely used in subsequent DPMs. 
Later, a brand new \textbf{ADM}~\cite{dhariwal2021diffusion} model emerges and leads the trend of guided diffusion models~\cite{liu2021more,nichol2021glide}
After that, a series of large DPMs~\cite{saharia2022photorealistic,rombach2022high,yu2022scaling,ramesh2022hierarchical} with billions of parameters have been proposed and have attracted the attention of a large number of researchers.\\

\noindent\textbf{Masked Auto-Encoders (MAEs)} Different from DPMs' route, as a series of efficient self-supervised visual learners, MAEs~\cite{li2021mst,zhou2021image,ma2022unitranser,zhang2022hivit,ma2022cmal} commit to pre-train a generalized representation models by \emph{mask-then-predict} pixels. Among them, \textbf{MAE}~\cite{he2022masked} is one of the most representative models, which proposes an asymmetric encoder-decoder architecture to feed those visible patches (about 25\%) into encoder and reconstructs the image by predicting the remaining 75\% patches. Subsequently, \textbf{SimMIM} proposes a simpler framework without the special designs (e.g., \emph{block-wise masking} and \emph{tokenization} via discrete VAE) to perform masked image modeling for addressing the data-hungry issue faced by large-scale model training.
Moreover, to train the hierarchical models faster and reduce the GPU memory consumption, \textbf{GreenMIM}~\cite{huang2022green} designs an optimal grouping strategy based on dynamic programming and couplings it with sparse convolution into MAEs, which enjoys a training-speed advantage in hierarchical ViT training, such as Swin Transformer~\cite{liu2021swin} and Twins Transformer~\cite{chu2021twins}. \\

\noindent\textbf{Compress-based Models.} Compress-based image reconstruction models~\cite{ramesh2021zero,vahdat2021score,kim2022maximum} aim to compress image into a smaller latent space for training acceleration, which is another line of research relevant to our work. 
\textbf{VQVAE}~\cite{van2017neural} proposes a simple yet powerful generative model to learn latent discrete representations by introducing vector quantization operation into VAEs and has shown powerful image compression capabilities. 
Based on the technique, \textbf{VQGAN}~\cite{esser2021taming} further models images as a composition of perceptually rich image constituents and introduces adversarial training for better image reconstruction. Recently, \textbf{Stable Diffusion}~\cite{rombach2022high} has become one of the most sought-after diffusion models among researchers, due to its excellent text-to-image synthesis performance. 
Though achieving remarkable progress, these compress-based models are basically built on top of the DPMs and therefore still suffer from the inherently strong temporal-dependence of the Markov diffusion, e.g., Stable Diffusion is still very computationally expensive if trained from scratch.

Motivated by these works, our \textbf{LMD} model focuses on unifying the  \underline{\textbf{L}}atent space project technique, the \underline{\textbf{M}}ask self-supervised technique and the \underline{\textbf{D}}iffusion generative idea, then respectively leverages these techniques to reduce the dimension of the input image, avoid temporal-dependence for parallel computing, and enhance spatial-dependence for faster training, which eventually reduces the total training time-consumption for faster image reconstruction. 

\section{Methodology}
To lower the computational demands and training time-consumption towards high-resolution image reconstruction, we propose a novel latent masking diffusion (LMD) framework, which integrates progressive mask self-supervised strategies into an encoder-decoder framework, as depicted in Figure~\ref{fig: model}. 
LMD mainly contains two steps: 1) Perceptual Latent Space Projection (Sec.~\ref{subsec:PLSP}) and 2) Latent Space Masking Diffusion (Sec.~\ref{subsec:LSMD}).  The former aims to pretrain a VAE-based latent space projector to compress input images into a perceptually high-dimensional space for acceleration, while the latter aims to conduct latent masking diffusion procedure for more efficient image reconstruction.

We observe that the mask autoencoders provide high GPU parallel computing efficiency due to using ViT as the visual learner, but there are still two problems: (i) Almost all existing MAEs are modeled at the pixel-level. Though the computing cost can be reduced by assigning a higher pixel proportion to each patch, it will greatly damage the global-perceptibility of semantic elements. (ii) The current MAEs follow a high-proportion masking strategy for training, which will greatly impact the spatial-dependence  among  pixels and make the training unstable and the time-consumption longer. A better way may be to gradually increase the mask proportion to make full use of the spatial-dependency for acceleration. We now introduce the details of the two steps in LMD.

\subsection{Perceptual Latent Space Projection}
\label{subsec:PLSP}

The latent space projection step is performed to compress the input images into a perceptual high-dimensional space by leveraging a pretrained latent space projector (LSP) based on previous work~\cite{esser2021taming,rombach2022high}. The LSP consists of an encoder $\mathcal{E}$, a decoder $\mathcal{G}$, a discriminator $\mathcal{D}$, and a learnable latent codebook $\mathcal{Z}$. Given an input image $x \in \mathbb{R}^{H\times W\times 3}$,  LSP first compress the image $x$ into a latent variable $\hat{z}$ by encoder $\mathcal{E}$, i.e., $\hat{z}=\mathcal{E}(x)$ and $\hat{z} \in \mathbb{R}^{h\times w\times d}$, where $h$ and $w$ respectively denote scaled height and width (scaled factor $f=H/h=W/w$), and $d$ is the dimensionality of the compressed latent variable. After going through the step described in Sec.~\ref{subsec:LSMD}, the latent variable $\hat{z}$ is updated and finally reconstructed into $\hat{x}$ by decoder $\mathcal{G}$. Formally, 
\begin{equation}
	\hat{x}=\mathcal{G}_{\theta}(\text{LSMD}_{\phi}(\mathcal{E}_{\theta}(x))),
\end{equation}
where LSMD($\cdot$) represents subsequent latent space masking diffusion step, $\phi$ denotes the parameters of LSMD, and $\theta$ denotes the parameters of LSP that are first pretrained and then frozen to use in the LSMD stage. 

\begin{figure}[t]
	\centering
	\includegraphics[width=1\linewidth]{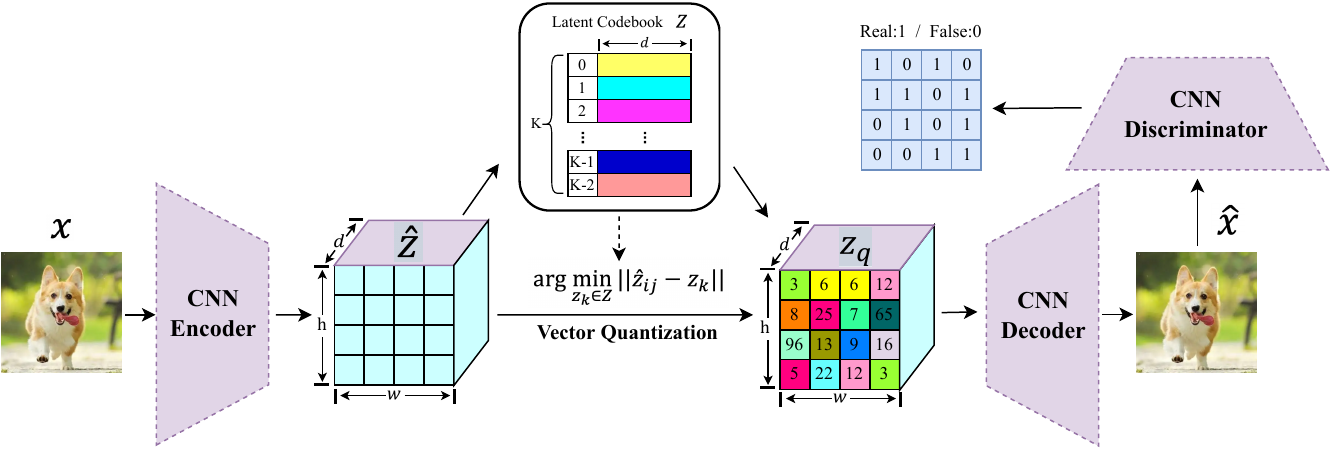}
        \vspace{-15pt}
	\caption{The Latent Space Projector.}
	\label{fig: latentProjector}
	\vspace{-10pt}
\end{figure}

\paragraph{Vector Quantization} As illustrated in Figure~\ref{fig: latentProjector},  vector quantization  in the pre-training stage of LSP aims to map the aforementioned compressed latent variable $\hat{z}$ into a perceptual latent variable $z_q$. A learnable codebook $\mathcal{Z}=\{z_{k}\}_{k=1}^K\subset \mathbb{R}^{d}$, which has been pre-trained by~\cite{van2017neural}, is introduced to help LSP  learn the perceptual latent feature of the image constituents.

Specifically, the codebook can be viewed as a discrete latent space of size $K$ that is leveraged to express the relatively complete and perceptual semantic elements of the original image constituents (e.g., dog's eyes or tongue), and can be retrieved by $\hat{z}$ to obtain a latent variable $z_{q}$ using element-wise vector quantization function \textbf{q}($\cdot$), i.e., $z_{q}=\textbf{q}(\hat{z})$, which is the key to compressing an image with little loss of accuracy. More precisely, the vector quantization function \textbf{q}($\cdot$) aims to obtain a perceptual discrete representation $z_{q}$ of the input image by using  $\hat{z}$ to perform nearest neighbour look-up over latent codebook $\mathcal{Z}$,  as follows,
\begin{equation}
	\label{Eq:q}
	\textbf{q}(\hat{z}):=\left(\arg\min_{z_{k}\in \mathcal{Z}}||\hat{z}_{ij}-z_{k}||\right)\in \mathbb{R}^{h\times w\times d}.
\end{equation}
In sum, the above forward process in the LSP stage can be formally described as,
\begin{equation}
	\hat{x}=\mathcal{G}_{\theta}(\textbf{q}_{\theta}(\mathcal{E}_{\theta}(x))).
\end{equation}
Note the operations in Formula~(\ref{Eq:q}) are non-differentiable, so the above forward process cannot be directly optimized due to the gradient of the \textbf{q}($\cdot$) cannot be backpropagated. Following~\cite{esser2021taming}, we adopt a straight-through gradient estimator to copy the gradients from the decoder to the encoder for end-to-end training via the loss function $\mathcal{L}_{\text{VQ}}$:
\begin{equation}
	\begin{aligned}
		\mathcal{L}_{\text{VQ}}(\mathcal{E},\mathcal{G},\mathcal{Z})=||x-\hat{x}||^2&+\ \ \  ||\text{sg}[\mathcal{E}(x)]-z_q||_2^2\\
		&+\beta||\text{sg}[z_q]-\mathcal{E}(x)]||_2^2,
	\end{aligned}
\end{equation}
where the ﬁrst term is reconstruction loss between $x$ and $\hat{x}$, the middle term is vector quantization loss between encoded vector $\hat{z}$ and retrieved vector $z_q$, here sg[$\cdot$] stands for stop-gradient operator and is used to solely update the $z_q$ part of the codebook. The last term is the commitment loss designed to ensure the encoder commits to an embedding and its output does not grow. To keep consistent with~\cite{van2017neural}, the commit factor $\beta$ is set to $0.25$.

\paragraph{Adversarial training} To make the LSP more robust, adversarial training is also introduced into our work. As mentioned above, the decoder $\mathcal{G}$ is leveraged to reconstruct the latent variable $z_q$ into $\hat{x}$.  After that, a patch-based discriminator is introduced to accept the real image patch from $x$ or the reconstructed image patch from $\hat{x}$ and give out a $(1, 0)$ judgment, which is trained via an adversarial loss $\mathcal{L}_{\text{ADV}}$:
\begin{equation}
	\mathcal{L}_\text{ADV}(\{\mathcal{E},\mathcal{Z},\mathcal{G}\},\mathcal{D})=-[log(1-\mathcal{D}(\hat{x}))+\log\mathcal{D}(x)).
\end{equation}
The total objective for ﬁnding the optimal latent projector is:
\begin{equation}
	\mathcal{L}_\text{Total}=\mathcal{L}_{\text{VQ}}(\mathcal{E},\mathcal{G},\mathcal{Z})+\gamma\mathcal{L}_\text{ADV}(\{\mathcal{E},\mathcal{Z},\mathcal{G}\},\mathcal{D}),
\end{equation}
where $\gamma$ is an adaptive weight hyperparameter. With the patch-level adversarial training,  LSP can be well generalized to the subsequent LSMD step, which is also trained on the same patch-level.

\subsection{Latent Space Masking Diffusion}
\label{subsec:LSMD}

The latent space masking diffusion step (i.e., LSMD$_{\phi}$($\cdot$)) follows an encoder-decoder architecture and is designed to achieve progressive masking diffusion by the following three components. 

\subsubsection{3.2.1\quad Latent Encoder}
\label{subsubsec:Encoder} 
\paragraph{}\vspace{-5pt}To learn the deeper semantics of the compressed latent variables $\hat{z}$ of the image $x$, we sequentially perform a pipeline, which includes patch embedding layer, mask scheduling layer, spatial position embedding layer and MAE encoder blocks, for training.

\begin{figure}[t]
	\centering
	\includegraphics[width=1\linewidth]{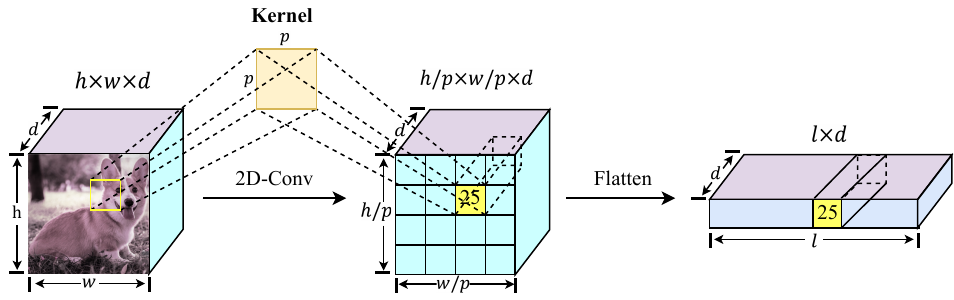}
        \vspace{-15pt}
	\caption{The Patch Embedding Layer.}
	\label{fig: PEL}
\vspace{-15pt}
\end{figure}

\paragraph{Patch Embedding Layer}~The patch embedding layer aims to further embed the compressed latent variable $\hat{z} \in \mathbb{R}^{h\times w\times d}$ into patched-based latent variable $\hat{z}_p\in \mathbb{R}^{l\times d}$ ($l$ is the number of the patches, and the size of each patch is $p\times p$ pixels), as illustrated in Figure~\ref{fig: PEL}. 

Specifically, we first feed the compressed latent variable $\hat{z}$ into a 2D-convolution layer to perform the convolution operation, where the kernel size and stride are both set to $p$, and then obtain the final patch-based representation $\hat{z}_p$ through a flatten operation. Note that $\hat{z}_p$ is a latent vector with length $l$, which is treated as the representation of a visual sequence and served as input to the subsequent visual Transformer.

\paragraph{Mask Scheduling Layer}~The mask scheduling layer is designed to produce an increasing mask-ratio sequence by using a masking diffusion scheduler, which will be detailed in Sec.~\ref{subsubsec:MDS}. Based such a mask-ratio sequence, LMD will progressively increase the 
mask  proportion of patches  in $\hat{z}_p$ with the increase of the training step, and finally only read out the unmasked patches for encoding.

\paragraph{Spatial Position Embedding Layer} Since the self-attention mechanism in Transformer is not sensitive to position, and the 1D position-features of tokens in NLP are ineffective for 2D data of image, we propose to use a spatial position embedding layer to learn the 2D-features and integrate them into latent variable $\hat{z}_p$ for obtaining better spatial-aware latent vector representations. Specifically, we first convert the index of each patch in $\hat{z}_p$ (see Figure~\ref{fig: model}) into their 2D coordinates ($c_x$, $c_y$) by dividing the index number into $h/p$ (or $w/p$) to get the quotient as $c_x$ and the remainder as $c_y$. Here $c_x$ and $c_y$  both are integers and record spatial position of the current patch. Then, we respectively embed them into a $\sin\raisebox{0mm}{-}\cos$ space via SPE($\cdot$),
\begin{equation}
	\text{SPE}(i,j)=\left\{\begin{array}{rr}
		\sin(\frac{i}{10000^{\frac{2i}{d}}}), & \text{if $i$ is even}  \\
		\cos(\frac{i}{10000^{\frac{2i-2}{d}}}), & \text { otherwise }
	\end{array}\right.
\end{equation}
to obtain a 2D position vector ($\hat{z}_x$, $\hat{z}_y$) corresponding to latent variable $\hat{z}_p$, where $i\in (1,l)$ and $j\in (1,d/2)$ respectively denotes patch and dimension index. 
Note here $\hat{z}_x \in \mathbb{R}^{\frac{d}{2}}$ and $\hat{z}_y\in \mathbb{R}^{\frac{d}{2}}$.  Finally, we can obtain the updated spatial-aware vector $\hat{z}_p \in \mathbb{R}^{d}$, by first concatenating $\hat{z}_x$ and $\hat{z}_y$ and then adding it to previous $\hat{z}_p$.
\paragraph{MAE Encoder Blocks} This layer is adopted to learn attentive representations for better mining the deeper sematics of compressed images. Similar to ViT-base, we adopt 12 layers of Transformer as MAE encoder blocks, and define each Transformer block by a block function $f_{\text{block}}(\cdot)$ as, 
\begin{equation}
	\hat{z}^{(\ell)}_{\text{out}}=f^{(\ell)}_{\text{block}}(\hat{z}^{(\ell-1)}_{\text{out}})
\end{equation}
\begin{equation}
	\hat{z}^{(0)}_{\text{out}}=\hat{z}_{p}
\end{equation}
where $\ell$ is layer index, and $\hat{z}^{(12)}_{\text{out}}$ is the output of the last layer. Note that in $\hat{z}_p$, only unmasked patches are fed into the above MAE encoder blocks for image reconstruction.

\begin{figure}[t]
	\centering
	\includegraphics[width=1\linewidth]{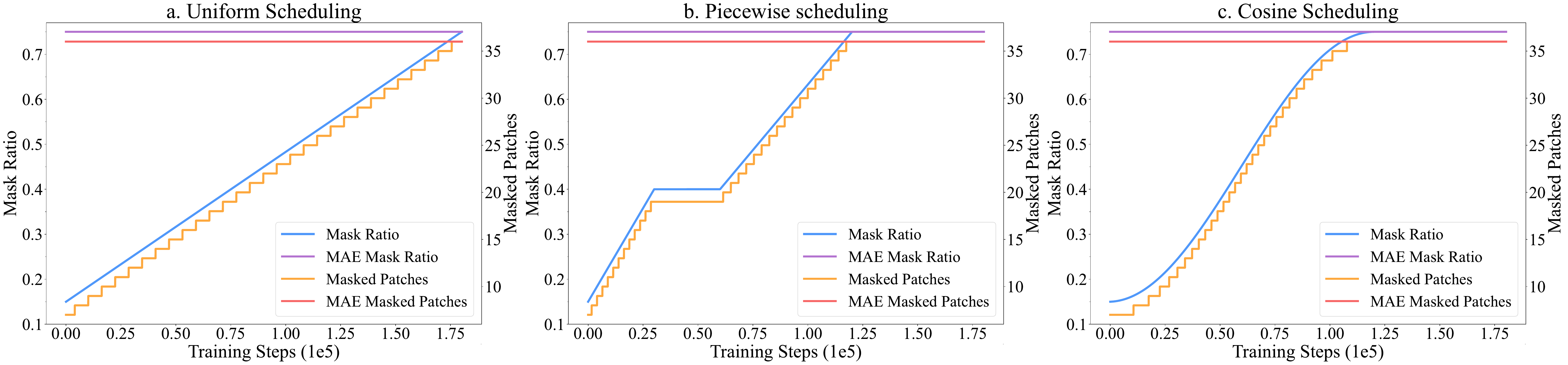}
\vspace{-15pt}
	\caption{The three masking diffusion scheduling schemes.}
	\label{fig: MDS}
\vspace{-10pt}
\end{figure}

\subsubsection{3.2.2\quad Latent Decoder}
\label{subsubsec:Decoder}
\paragraph{}\vspace{-5pt}As opposed to the above encoder, the decoder is designed to reconstruct the compressed latent variable $\hat{z}\in \mathbb{R}^{h\times w\times d}$ by using the output $\hat{z}_\text{out}$ of the encoder as input. The pipeline of the decoder consists of a latent linear layer, a spatial position recall layer, MAE decoder blocks and latent normalization and prediction layers, and is also executed sequentially. Wherein, the latent linear layer and asymmetric MAE decoder blocks are adopted to given extra consideration for trade off between discriminant and generative tasks, the spatial position recall layer keeps the same operation as in the encoder but with negative SPE($\cdot$), and then the normalization and prediction layer is used for latent image reconstruction.

\paragraph{Trade off between discriminant tasks and generative tasks} In the original MAE~\cite{he2022masked}, the asymmetric structure is adopted by setting up a heavier encoder (e.g., \emph{12 blocks}) and a lighter decoder (e.g., \emph{8 blocks}), which is more suitable for  discriminant tasks. We consider to take a trade off between discriminant tasks and generative tasks, and additionally propose to use a lighter encoder (e.g., \emph{8 blocks}) and a heavier decoder (e.g., \emph{12 blocks}) for generative tasks such as image synthesis. Moreover, a latent linear layer is also preferentially placed at the beginning of the decoder for deeper decoder embedding (e.g., embed \emph{512-dim} to \emph{1024-dim} for decoding).

\paragraph{Latent Image Reconstruction} Different from the previous MAEs~\cite{he2022masked,xie2022simmim}, the latent image reconstruction (LIR) target of LMD is established in a latent space. The last layer of the decoder is a latent linear prediction layer whose number of output channels equals the dimension $d$ of $\hat{z}$, which is used to predict the reconstructed latent image $\hat{z}_\text{rec}$. LMD adopts the mean squared error (MSE) between the reconstructed latent variable $\hat{z}_{\text{rec}}$ and original latent variable $\hat{z}$ of the compressed image as training target,
\begin{equation}
	\label{formula:LIR}
	\mathcal{L}_{\text{LIR},\phi}=||\hat{z}-\hat{z}_\text{rec}||^2_2,
\end{equation}
here $\phi$ denote the parameters of the whole LSMD step. Similar to MAE, the loss is only computed on masked patches.

\setlength{\tabcolsep}{3pt}
\begin{table*}[]
	\small
	\begin{tabular}{llcccccc}
		\hline\hline
		\multicolumn{8}{c}{\textbf{ImageNet-1K (IN1K)}} \\ \hline
		\multicolumn{1}{l|}{\textbf{Method}} &
		\multicolumn{1}{c|}{\textbf{Backbone}} &
		\multicolumn{1}{c|}{\textbf{Image Size}} &
		\multicolumn{1}{c|}{\textbf{Patch Size}} &
		\multicolumn{1}{c|}{\textbf{Mask Ratio}} &
		\multicolumn{1}{c|}{\textbf{MIT}\color{red}$\downarrow$} &
		\multicolumn{1}{c|}{\textbf{MLT}\color{red}$\downarrow$} &
		\textbf{MLI}\color{red}$\downarrow$\\ \hline\hline
		\multicolumn{1}{l|}{\textbf{MAE}~(\citeyear{he2022masked}) } &
		\multicolumn{1}{l|}{ViT-B (12 encoder blocks + 8 decoder blocks)} &
		\multicolumn{1}{c|}{224 x 224} &
		\multicolumn{1}{c|}{16 x 16} &
		\multicolumn{1}{c|}{0.75} &
		\multicolumn{1}{c|}{2.62} &
		\multicolumn{1}{c|}{17.11} &
		6.53 \\
		\multicolumn{1}{l|}{\textbf{SimMIM}~(\citeyear{xie2022simmim})} &
		\multicolumn{1}{l|}{Swin-B (hierarchical 2, 2, 18 and 2 blocks)} &
		\multicolumn{1}{c|}{192 x 192} &
		\multicolumn{1}{c|}{32 x 32} &
		\multicolumn{1}{c|}{0.6} &
		\multicolumn{1}{c|}{3.45} &
		\multicolumn{1}{c|}{20.01} &
		5.80 \\
		\multicolumn{1}{l|}{\textbf{GreenMIM}~(\citeyear{huang2022green})} &
		\multicolumn{1}{l|}{Swin-B (hierarchical 2, 2, 18 and 2 blocks)} &
		\multicolumn{1}{c|}{224 x 224} &
		\multicolumn{1}{c|}{4 x 4} &
		\multicolumn{1}{c|}{0.75} &
		\multicolumn{1}{c|}{2.23} &
		\multicolumn{1}{c|}{11.93} &
		5.35 \\
		\multicolumn{1}{l|}{\textbf{UM-MAE}~(\citeyear{li2022uniform})} &
		\multicolumn{1}{l|}{ViT-B (12 encoder blocks + 8 decoder blocks)} &
		\multicolumn{1}{c|}{256 x 256} &
		\multicolumn{1}{c|}{16 x 16} &
		\multicolumn{1}{c|}{0.25} &
		\multicolumn{1}{c|}{\textbf{2.15}} &
		\multicolumn{1}{c|}{13.80} &
		6.42 \\ \hline
		\multicolumn{1}{l|}{\textbf{LMD-PS (Ours)}} &
		\multicolumn{1}{l|}{ViT-B (12 encoder blocks + 8 decoder blocks)} &
		\multicolumn{1}{c|}{224 x 224} &
		\multicolumn{1}{c|}{16 x 16} &
		\multicolumn{1}{c|}{0.15 / 0.4 / 0.75} &
		\multicolumn{1}{c|}{2.78} &
		\multicolumn{1}{c|}{8.37} &
		3.01\\
		\multicolumn{1}{l|}{\textbf{LMD-CS (Ours)}} &
		\multicolumn{1}{l|}{ViT-B (12 encoder blocks + 8 decoder blocks)} &
		\multicolumn{1}{c|}{224 x 224} &
		\multicolumn{1}{c|}{16 x 16} &
		\multicolumn{1}{c|}{cosine-based values} &
		\multicolumn{1}{c|}{2.61} &
		\multicolumn{1}{c|}{\textbf{6.92}} &
		\textbf{2.65} \\ \hline\hline
	\end{tabular}
        \vspace{-5pt}
	\caption{The pre-training mean time-consumption versus MAEs methods on ImageNet-1K dataset.}
        \vspace{-5pt}
	\label{table: maes_results}
\end{table*}

\subsubsection{3.2.3\quad Masking Diffusion Scheduler}   
\label{subsubsec:MDS}

\begin{figure}[t]
	\centering
	\includegraphics[width=1\linewidth]{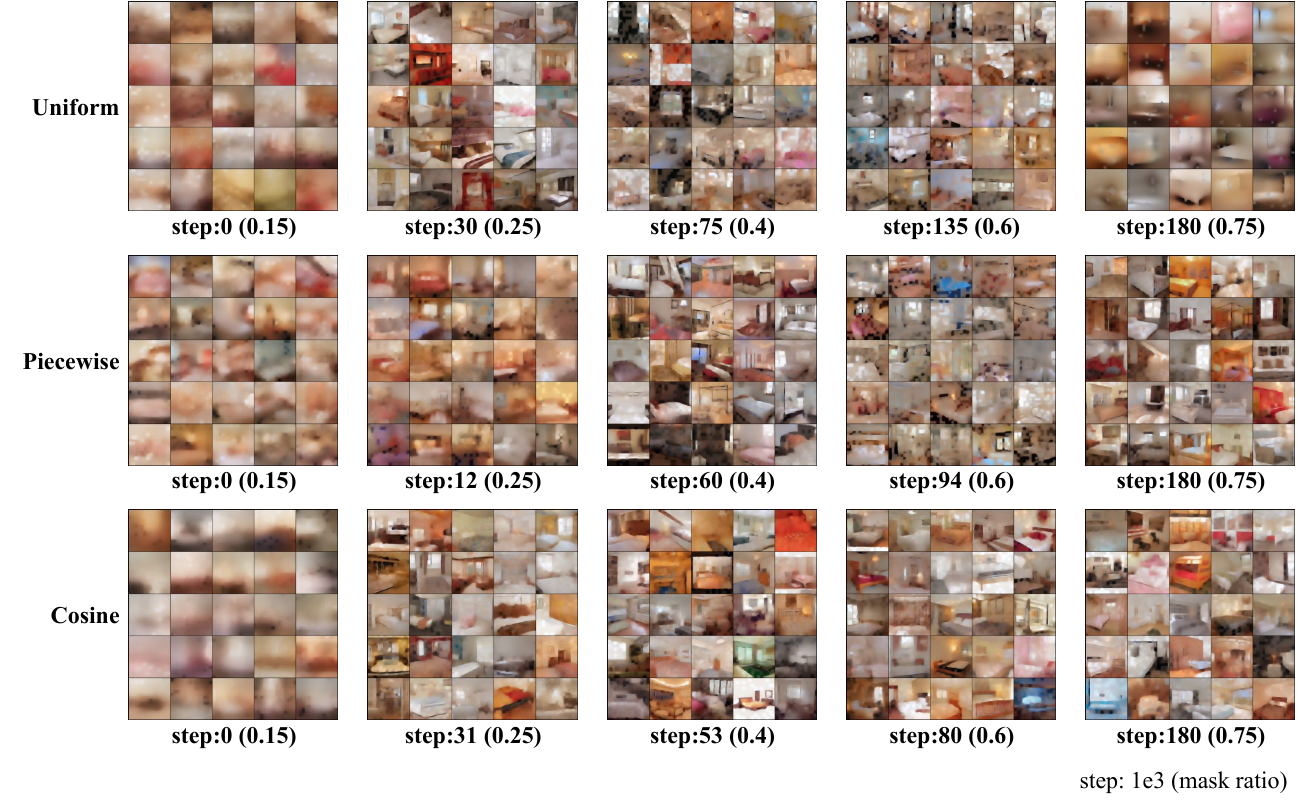}
	\vspace{-15pt}
	\caption{The masking diffusion samples (The 5$\times$5 images from the LSUN-bedrooms dataset are sampled only at a maximum of 180k training step in latent space).}
	\label{fig: MDSamples}
	\vspace{-10pt}
\end{figure}

\paragraph{}\vspace{-5pt}The masking diffusion scheduler is the key to LMD, which aims to produce an increasing mask-ratio sequence for dynamically fitting the model training, so as to optimize the overall training time-consumption. Motivated by the lower temporal-dependence of MAEs and higher spatial-dependence of DPMs, we stand on the shoulder of mask self-supervise technique and diffusion generation mode to propose the masking diffusion strategy, which is achieved through the following three scheduling schemes\footnote{In all schemes, the mask ratio of randomly masking is preset in $[0.15\raisebox{0mm}{-}0.75]$, which follows the discrete optimal lower bound in BERT~\cite{kenton2019bert} and the continuous optimal upper bound in MAE~\cite{he2022masked}.}.

\paragraph{Uniform Scheduling} This uniform scheduling (Figure~\ref{fig: MDS}a) is a pre-explored scheme, which follows the assumption that the difficulty of model training decreases uniformly with the increase of the number of training steps. In the training stage, we randomly sampled $5\times 5$ latent images for evaluation, as shown in Figure~\ref{fig: MDSamples}. From Figure~\ref{fig: MDSamples}, we can observe that when the mask ratio grows to $0.4$, the masking speed of the uniform scheduler exceeds the reconstructing speed of the model. Moreover, it can be noticed that when the mask ratio reaches  $0.75$, this phenomenon is further aggravated. 

\paragraph{Piecewise Scheduling} The previous pre-explored scheme shows that the model capability is not uniformly improved with the growth of the mask ratio. 
To the end, we provide the piecewise scheduling scheme, as shown in Figure~\ref{fig: MDS}b. Specifically, for the first 1/6 training steps, we increase the mask ratio linearly  from $0.15$ and $0.4$, for the next 1/6 training steps, we maintain the mask ratio at $0.4$, and for the remaining 1/3 steps, we continue to linearly increase until reaches $0.75$, and maintain such a ratio till the end of the training.

\paragraph{Cosine Scheduling} From Figure~\ref{fig: MDSamples}, we can observe that under piecewise scheduling, the model  still performs poor in both the early stages (e.g., at step $12 e^3$) and the later stages (e.g., at step $94 e^3$). We guess the reasons why the model requires more training steps in the early and later stages are respectively due to poor data fitting  and high mask ratio. 
To address this issue, we further propose a new cosine scheduling scheme. Specifically, considering that the cosine function can approximate the scheduling process well, we adopt the cosine function to compute a sequence of mask ratios. As can been seen from Figure~\ref{fig: MDSamples}, such a scheduling scheme achieves the best results.

\section{Experiments}


\setlength{\tabcolsep}{3pt}
\begin{table}[]
	\small
	\begin{tabular}{l|c|c|c}
		\hline\hline
		\textbf{Method}        &\textbf{Mask Ratio}    &  \textbf{MAT@1} \color{red}$\downarrow$ & \textbf{MAT@5} \color{red}$\downarrow$ \\ \hline\hline
		\textbf{MAE}~(\citeyear{he2022masked})           & 0.75          &  0.374 & 0.253\\
		\textbf{LMD-PS (Ours)} & 0.15 / 0.4 / 0.75 &  0.158     & 0.137    \\
		\textbf{LMD-CS (Ours)} & cosine-based values       &  \textbf{0.135}     & \textbf{0.102}    \\ \hline\hline
	\end{tabular}
        \vspace{-5pt}
	\caption{The fine-tuning valid mean time-consumption for accuracy improvement on ImageNet-1K dataset.}
        \vspace{-10pt}
	\label{table: maes_ft_results}
\end{table}

\setlength{\tabcolsep}{2pt}
\begin{table}[]
	\footnotesize
	\begin{tabular}{lccccc}
		\hline\hline
		\multicolumn{6}{c}{\textbf{LSUN-Bedrooms 256 x 256}}                                                                                                 \\ \hline
		\multicolumn{1}{l|}{\textbf{Method}} & \multicolumn{1}{c|}{\textbf{Backbone}} & \multicolumn{1}{c|}{\textbf{Diffusion Steps}} & \multicolumn{1}{c|}{\textbf{MIT}\color{red}$\downarrow$}   & \multicolumn{1}{c|}{\textbf{MLT}\color{red}$\downarrow$}   & \textbf{MLI}\color{red}$\downarrow$   \\ \hline\hline
		\multicolumn{1}{l|}{\textbf{DDPM}~(\citeyear{ho2020denoising})}   & \multicolumn{1}{c|}{U-Net}    & \multicolumn{1}{c|}{1000}            & \multicolumn{1}{c|}{5.45} & \multicolumn{1}{c|}{39.89} & 7.32 \\
		\multicolumn{1}{l|}{\textbf{iDDPM}~(\citeyear{nichol2021improved})}      & \multicolumn{1}{c|}{U-Net} & \multicolumn{1}{c|}{1000} & \multicolumn{1}{c|}{5.32} & \multicolumn{1}{c|}{24.74} & 4.65 \\
		\multicolumn{1}{l|}{\textbf{DDIM}~(\citeyear{song2020denoising})}       & \multicolumn{1}{c|}{U-Net} & \multicolumn{1}{c|}{1000} & \multicolumn{1}{c|}{4.46} & \multicolumn{1}{c|}{29.53} & 6.62 \\
		\multicolumn{1}{l|}{\textbf{LDM}~(\citeyear{rombach2022high})}        & \multicolumn{1}{c|}{U-Net} & \multicolumn{1}{c|}{1000} & \multicolumn{1}{c|}{2.86} & \multicolumn{1}{c|}{15.84} & 5.54 \\ \hline
		\multicolumn{1}{l|}{\textbf{LMD (Ours)}} & \multicolumn{1}{c|}{ViT-B} & \multicolumn{1}{c|}{29}    & \multicolumn{1}{c|}{\textbf{2.45}} & \multicolumn{1}{c|}{\textbf{7.06}} & \textbf{2.88} \\ \hline\hline
	\end{tabular}
        \vspace{-5pt}
	\caption{The training mean time-consumption versus DPMs methods on LSUN-Bedrooms dataset.}
	\label{table: dpms_results}
	\vspace{-20pt}
\end{table}

\subsection{Experimental Setup}
\noindent\textbf{Datasets and Metrics.} Following~\cite{he2022masked,ho2020denoising}, we pre-train our model on ImageNet-1K (IN1K)~\cite{deng2009imagenet} and LSUN-Bedrooms~\cite{yu2015lsun} respectively. Three main  metrics: mean iteration time (MIT), mean loss-decrease time (MLT) and mean loss-decrease iterations (MLI) are adopted to evaluate the valid training time-consumption of the models, which will be detailed in the Appendix. On the downstream classification task, we evaluate our model on the IN1K dataset (with 1000 object categories) and adopt mean accuracy-increase time (MAT) including MAT@1 and MAT@5 as the main evaluation metrics, which can be used to test the valid time-consumption with mean accuracy improvement. Moreover, FID, CLIP-score and LPIPS metrics are used to evaluate the generative performance of our models and some qualitative cases generated by LMD will be presented in the appendix.

\vspace{8pt}
\noindent\textbf{Baselines.}~Our model can be regarded as the unification of ``Latent + Mask + Diffusion'' and can be generalized to discriminant and generative tasks by a simple trade-off. Therefore, for a more holistic comparison, we compare LMD with the two categories of baseline models: 1) DPMs, including DDPM~\cite{ho2020denoising}, iDDPM~\cite{nichol2021improved}, DDIM~\cite{song2020denoising} and LDM~\cite{rombach2022high}; 2) MAEs, including MAE~\cite{he2022masked}, SimMIM~\cite{xie2022simmim}, GreenMIM~\cite{xie2022simmim} and UM-MAE~\cite{li2022uniform}. For a fair comparison, all baselines use the \emph{base} model.

\vspace{8pt}
\noindent\textbf{Implementation Details.} 
 LMD adopts $20$-layers ViT as the backbone, of which $8$ encoder blocks and $12$ decoder blocks for generative training, and $12$ encoder blocks and $8$ decoder blocks for discriminant training. The mask scheduling scheme is set to Cosine scheduling. The mask ratio of the mask scheduler is set in $[0.15, 0.75]$. The scaling factor $f$ is set as $8$. The base learning rate is set as $1.5e^{-4}$, and the weight decay is set as $0.05$. We use the Adan~\cite{xie2022adan} optimizer to optimize the model and all our experiments are performed on $2$ NVIDIA RTX3090 GPUs with PyTorch framework. 

%
%

\subsection{Overall Performance}
As shown in Table~\ref{table: maes_results}, compared with MAEs methods, LMD achieves the best performance on MLT and MLI metrics. As for the MIT metric, although UM-MAE achieves the optimal mean iteration time, but with a very low mask ratio  of $0.25$.  In fact, the MLT metric can better reflect the efficiency of the model for valid loss decreases, and we are nearly twice as fast as UM-MAE on this metric. From Table~\ref{table: maes_ft_results}, we can observe from this fine-tuning results that LMD is much faster than MAE, and the actual speed up is about 3$\times$ under the same accuracy contribution, which further proves LMD's effectiveness. Moreover, Table~\ref{table: dpms_results} shows the comparison between our model and DPMs-based methods in terms of training time. From Table~\ref{table: dpms_results}, it can be clearly seen that LMD greatly accelerate the generative training process, which proves that the LMD model is substantially more efficient in training as compared to the DPMs-based methods, because it can take full advantage of the parallel computing of  GPU by utilizing  ViT as the backbone. Further, Table~\ref{table: generative_results} shows that our LMD has achieved more competitive results in generative performance than SD-v1.4 or the recent Muse$_{base}$ with the best FID, CLIP-score and LPIPS results.

\setlength{\tabcolsep}{3pt}
\begin{table}[]
\centering
\small
\begin{tabular}{llll}
\hline\hline
\multicolumn{1}{l|}{}          & \multicolumn{1}{c|}{\textbf{FID}\color{red}$\downarrow$}   & \multicolumn{1}{l|}{\textbf{CLIP-score}\color{red}$\uparrow$} & \textbf{LPIPS}\color{red}$\downarrow$ \\ \hline\hline
\multicolumn{1}{l|}{\textbf{SD-v1.4}~\cite{rombach2022high}}   & \multicolumn{1}{c|}{17.01} & \multicolumn{1}{c|}{0.24}       & \multicolumn{1}{c}{0.45}  \\
\multicolumn{1}{l|}{\textbf{Muse}$_{base}$~\cite{chang2023muse}} & \multicolumn{1}{c|}{6.8}   & \multicolumn{1}{c|}{0.25}       & \multicolumn{1}{c}{0.33}  \\ 
\multicolumn{1}{l|}{\textbf{LMD-CS (Ours)}}      & \multicolumn{1}{c|}{\textbf{6.2}}   & \multicolumn{1}{c|}{\textbf{0.26}}       & \multicolumn{1}{c}{\textbf{0.27}}  \\ \hline\hline
                               &                            &                                 &      
\end{tabular}
\vspace{-15pt}
	\caption{Generative evaluation on CC~\cite{sharma2018conceptual}.}
	\label{table: generative_results}
	\vspace{-5pt}
\end{table}

\subsection{Ablation Studies}
In this part, we perform ablation experiments to evaluate the impact of each setting in our LMD on training time-consumption. We focus on three crucial settings, as shown in Table~\ref{table: ablation}. Specifically, $\#1$ indicates the complete LMD model; $\#2$ w/o LSP denotes that we remove the latent space projector and directly perform the masking diffusion strategy on the pixel space; $\#3$ w/o MAE denotes that we remove the MAE encoder-decoder blocks and replace with a U-Net~\cite{ronneberger2015u} model; $\#4$ w/o MDS denotes that we remove the masking diffusion scheduler and adopt a fixed 0.75 mask ratio for reconstruction. From Table~\ref{table: ablation}, we can observe that removing each component will result in a time-consumption increasing, which proves the effectiveness of all the settings employed by LMD. In particular, w/o MAE caused a mean delay of 2.70, while w/o MDS only caused a mean delay of 1.27, indicating that low temporal-dependence directly promotes the reduction of the training time-consumption, while high spatial-dependence is a secondary optimization factor, which indirectly reduces the training time-consumption. In contrast,  w/o LSP only caused an average delay of $0.33$. We suspect that this is due to the adoption of ViT to offset latent space acceleration, but it may also have a significant impact on reducing GPU memory consumption.

\subsection{Further Analysis}

\begin{figure}[t]
	\centering
	\includegraphics[width=1\linewidth]{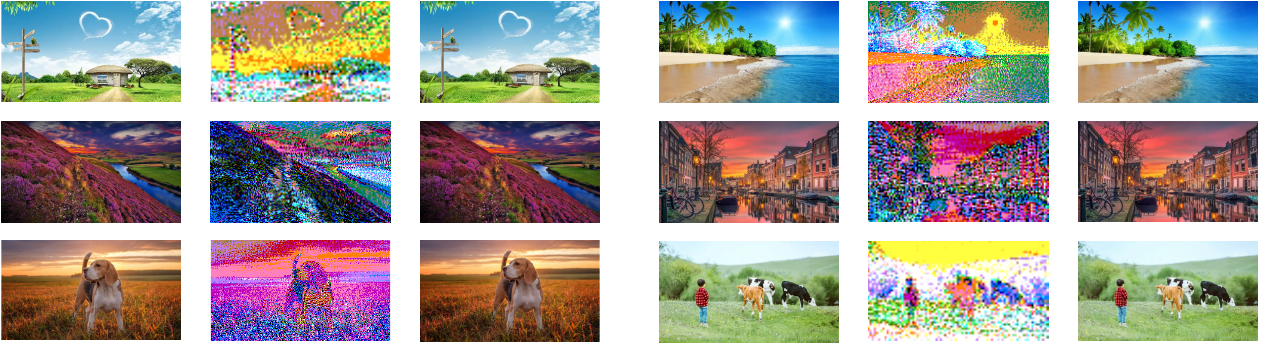}
	\vspace{-15pt}
	\caption{Case studies to test the impact of latent projector.}
	\label{fig: LPSsamples}
	\vspace{-15pt}
\end{figure}

\noindent\textbf{The impact of latent space projector.}~To better illustrate the effectiveness of our model in latent space reconstruction, we sample and visualize the latent image $\hat{z}$ (middle of the three columns in Figure~\ref{fig: LPSsamples}) compressed by the latent space projector and ultimately reconstructed images (right of the three columns  in Figure~\ref{fig: LPSsamples}). It can be observed that our latent space projector based on VQ-GAN almost achieves lossless image compression, so the accuracy of latent space image reconstruction can be guaranteed. Moreover, when the latent space projector fails on partial images (loss of accuracy due to unseen feature distributions), our model can ensure better generalization by adding an explicit reconstruction loss $||x-\hat{x}||^2_2$ to efficiently fine-tune the ViT blocks, without re-training the latent space projector. \\

\begin{figure}[t]
	\centering
	\subfloat[\scriptsize{LMD-PS.}]{
		\begin{minipage}[t]{0.22\textwidth}
			\flushleft
			\includegraphics[height=0.75\textwidth]{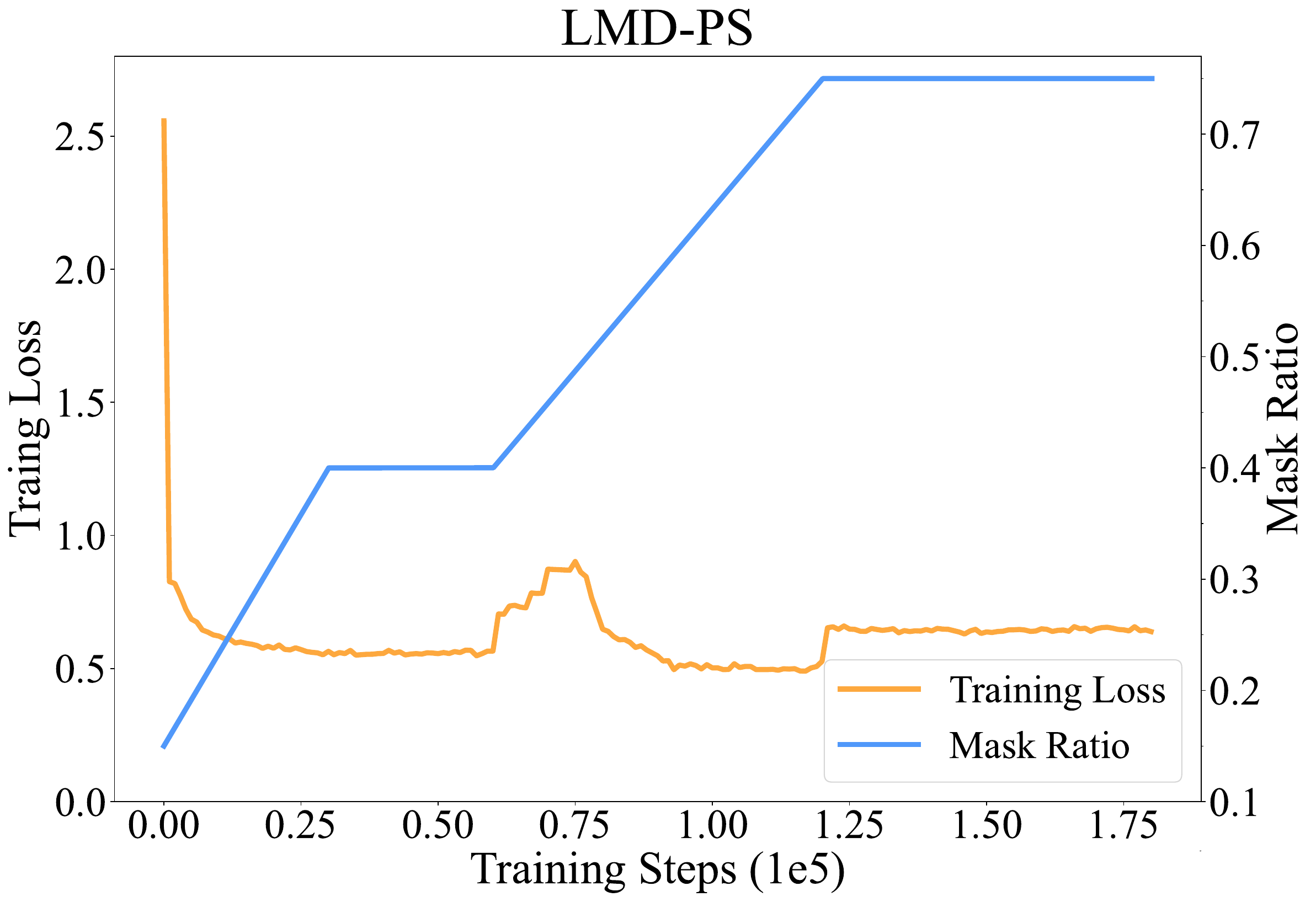}
			\vspace{-0.5em}
			\label{fig:LMD-PS}
		\end{minipage}
	}\hspace{4pt}
	\subfloat[\scriptsize{LMD-CS.}]{
		\begin{minipage}[t]{0.22\textwidth}
			\flushright
			\includegraphics[height=0.75\textwidth]{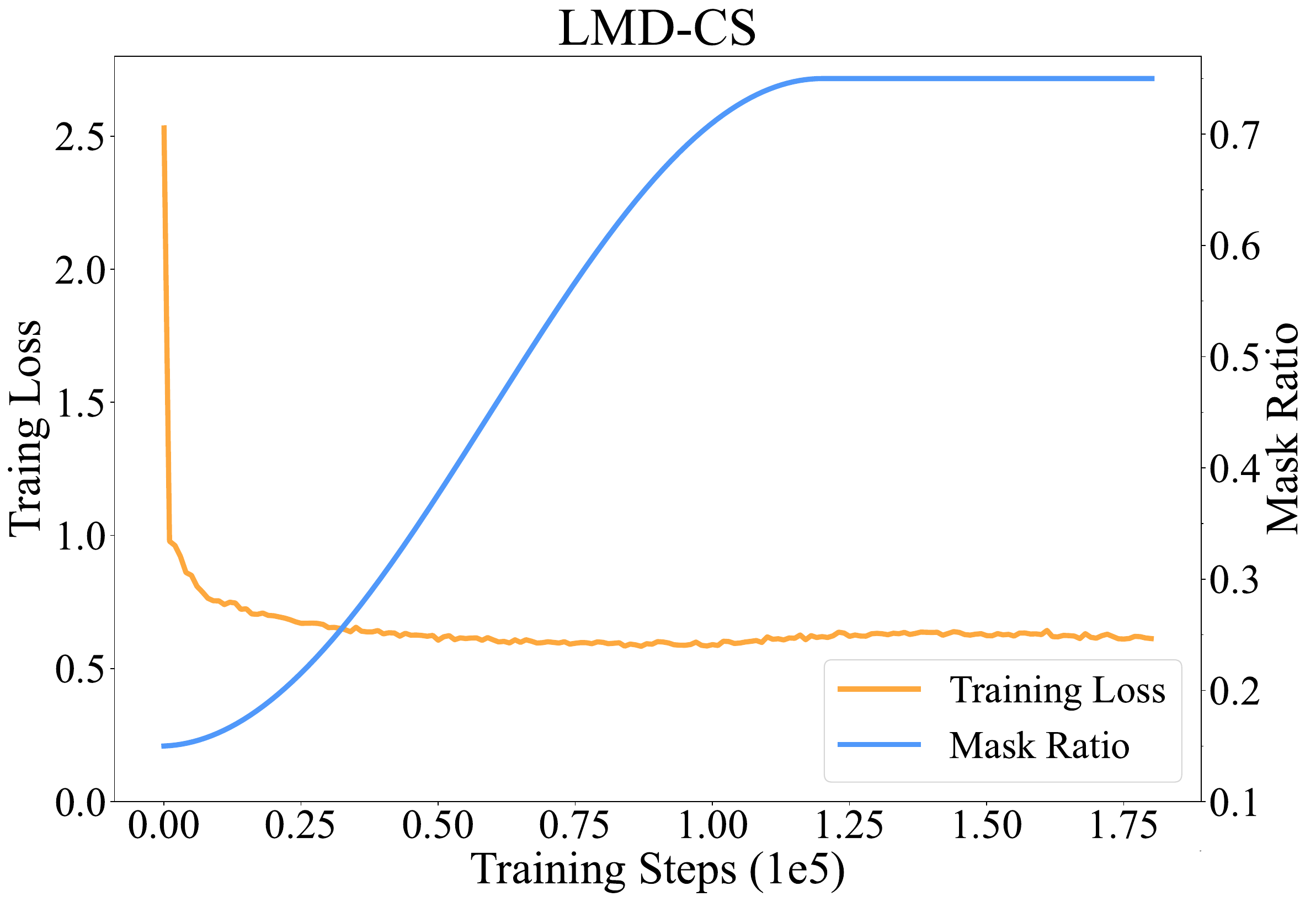}
			\vspace{-0.5em}
			\label{fig:LMD-CS}
		\end{minipage}
	}\\
	\vspace{-5pt}
	\caption{The loss curves from LMD-PS and LMD-CS.}
	\label{fig: MDS_loss}
	\vspace{-15pt}
\end{figure}

\noindent\textbf{Exploration of mask scheduling schemes.}~To more clearly illustrate the impact of piecewise scheduling and cosine scheduling on loss decline and show why cosine scheduling is more effective, we respectively visualize the loss decline curves from LMD-PS and LMD-CS, as shown in Figure~\ref{fig: MDS_loss}. From~Figure~\ref{fig: MDS_loss} (a), we can see that there is an obvious fluctuation when the mask ratio exceeds $0.4$, which indicates that the learning efficiency of the model at this time is slightly slower than that of the mask scheduler. This can be solved by two different ways: (1) adding more training steps at $0.4$ mask ratio; (2) adopting a more gentle scheduling scheme (such as cosine scheduling) within the range of $0$ to $0.4$. The former requires more training time to make up for the learning efficiency of the model (similar to MAEs with fixed mask ratio). In contrast, the cosine-based scheduling scheme shown in Figure~\ref{fig: MDS_loss} (b) is more gentle in the decline of losses, which proves its potential advantage in reducing training time-consumption.


\section{Conclusion}

\setlength{\tabcolsep}{1.5pt}
\begin{table}[]
	\footnotesize
	\begin{tabular}{l|ll|cc|l|l|l|c|c}
		\hline\hline
		\multirow{2}{*}{\#} &
		\multicolumn{2}{c|}{\multirow{2}{*}{\textbf{Setting}}} &
		\multicolumn{2}{c|}{\textbf{Overall}\color{red}$\downarrow$} &
		\multicolumn{1}{c|}{\textbf{MIT}} &
		\multicolumn{1}{c|}{\textbf{MLT}} &
		\multicolumn{1}{c|}{\textbf{MLI}} &
		\textbf{MAT@1} &
		\textbf{MAT@5} \\ \cline{4-10} 
		&
		\multicolumn{2}{c|}{} &
		\multicolumn{1}{c|}{\textbf{mean}} &
		\textbf{$\Delta$} &
		\multicolumn{1}{c|}{\textbf{train}} &
		\multicolumn{1}{c|}{\textbf{train}} &
		\multicolumn{1}{c|}{\textbf{train}} &
		\textbf{dev} &
		\textbf{dev} \\ \hline\hline
		1 & \multicolumn{2}{l|}{\textbf{LMD (Ours)}} & \multicolumn{1}{c|}{\textbf{2.48}} & -  & \textbf{2.61} & \textbf{6.92} & \textbf{2.65} & \textbf{0.135} & \textbf{0.102} \\ \hline
		2 & \multicolumn{1}{l|}{-L}  & w/o LSP  & \multicolumn{1}{c|}{2.81} & + 0.33 & 3.18 & 7.92 & 2.49 & 0.285 & 0.198 \\ \hline
		3 & \multicolumn{1}{l|}{-M}  & w/o MAE  & \multicolumn{1}{c|}{5.18} & +2.70 & 2.95 & 16.52 & 5.60 & 0.425 & 0.385 \\ \hline
		4 & \multicolumn{1}{l|}{-D}  & w/o MDS  & \multicolumn{1}{c|}{3.75} &+1.27  &2.62 &11.40  & 4.35 & 0.225 & 0.143 \\ \hline\hline
	\end{tabular}
        \vspace{-5pt}
	\caption{Ablation study on IN1K dataset.}
        \vspace{-5pt}
	\label{table: ablation}
\end{table}

In this paper,  we propose LMD, a latent masking diffusion framework for faster image synthesis. Specifically, we first employ a pre-trained latent space projector to compress the input image into a latent space with smaller scale to obtain their latent feature map, and then split the latent feature map into a patch sequence for masked self-supervised training. Unlike conventional using fixed mask ratio to reconstruction, we propose to gradually increase the masking ratio by mask schedulers and reconstruct the images by a progressive diffusion mode. By unifying the latent space project technique, mask self-supervised technique and diffusion generative scheme, LMD can respectively achieve the reduction the dimension of the input image, avoid temporal-dependence for parallel computing, and enhance spatial-dependence for fewer number of iterations, which is used to eventually reduce the total training time-consumption. Experiments on the representative ImageNet-1K and LSUN-Bedrooms datasets demonstrate the effectiveness and superior performance of  LMD, and illustrates its high-efficiency in training both generative and discriminant tasks.

\section{Acknowledgments}
We would like to thank all anonymous reviewers for their valuable comments. This work was partially supported by the Project funded by China Postdoctoral Science Foundation (No.2023M741950).

\bibliography{aaai24}

\end{document}